\documentclass{article}
\usepackage{ijcai23}

\usepackage{times}
\usepackage{soul}
\usepackage{url}
\usepackage[hidelinks]{hyperref}
\usepackage[utf8]{inputenc}
\usepackage[small]{caption}
\usepackage{graphicx}
\usepackage{amsmath}
\usepackage{amsthm}
\usepackage{booktabs}
\usepackage{algorithm}
\usepackage{algorithmic}
\usepackage[switch]{lineno}

\usepackage{bbm}
\usepackage{multirow}
\usepackage{amsfonts}
\usepackage{array}
\usepackage{subcaption}

\title{DenseLight: Efficient Control for Large-scale Traffic Signals with Dense Feedback}

\author{
Junfan Lin$^1$\and
Yuying Zhu$^1$\and
Lingbo Liu$^{2,3}$\and
Yang Liu$^{1}$\footnote{Corresponding author: Yang Liu (liuy856@mail.sysu.edu.cn)}\and
Guanbin Li$^1$\And
Liang Lin$^1$
\affiliations
$^1$The School of Computer Science and Engineering, Sun Yat-sen University, Guangzhou, China.\\
$^2$Department Land Surveying and Geo-Informatic, The Hong Kong Polytechnic University, Hongkong. \\
$^3$Smart Cities Research Institute, The Hong Kong Polytechnic University, Hong Kong
\emails
\{linjf8, zhuyy76\}@mail2.sysu.edu.cn,
lingbo.liu@polyu.edu.hk,
\{liuy856, liguanbin\}@mail.sysu.edu.cn,
linliang@ieee.org
}

\begin{document}

\maketitle
\begin{abstract}

Traffic Signal Control (TSC) aims to reduce the average travel time of vehicles in a road network, which in turn enhances fuel utilization efficiency, air quality, and road safety, benefiting society as a whole. Due to the complexity of long-horizon control and coordination, most prior TSC methods leverage deep reinforcement learning (RL) to search for a control policy and have witnessed great success. However, TSC still faces two significant challenges. 1) The travel time of a vehicle is delayed feedback on the effectiveness of TSC policy at each traffic intersection since it is obtained after the vehicle has left the road network. Although several heuristic reward functions have been proposed as substitutes for travel time, they are usually biased and not leading the policy to improve in the correct direction. 2) The traffic condition of each intersection is influenced by the non-local intersections since vehicles traverse multiple intersections over time. Therefore, the TSC agent is required to leverage both the local observation and the non-local traffic conditions to predict the long-horizontal traffic conditions of each intersection comprehensively. To address these challenges, we propose DenseLight, a novel RL-based TSC method that employs an unbiased reward function to provide dense feedback on policy effectiveness and a non-local enhanced TSC agent to better predict future traffic conditions for more precise traffic control. Extensive experiments and ablation studies demonstrate that DenseLight can consistently outperform advanced baselines on various road networks with diverse traffic flows. The code is available at \url{https://github.com/junfanlin/DenseLight}.

\end{abstract}

\section{Introduction}
Reducing traffic congestion is an essential task for efficient modern urban systems. As the number of vehicles in the cities increases year by year, backward traffic coordination not only does harm to the driving experience but also aggravates air contamination with more harmful fuel emissions~\cite{zhang2013air}. Alleviating traffic congestion by efficient traffic signal control (TSC)~\cite{mirchandani2001real} is one of the most practical and economical approaches~\cite{taylor2002rethinking}. Specifically, TSC aims at coordinating the traffic lights of a road network to regulate the traffic flows to minimize the average travel time of vehicles. However, the traffic data collected from a road network is usually massive yet incomprehensible \cite{liu2020dynamic,wang2023urban}. Therefore, the widely-adopted TSC strategies either fix signal routine~\cite{roess2004traffic} or adapt the traffic signal plans in real-time according to traffic flow patterns~\cite{cools2013self}.

To automatically mine useful information from the massive traffic data, more and more studies leverage the powerful representation capability of deep neural networks~\cite{lecun2015deep} to learn TSC agents~\cite{van2016coordinated,zhang2021expression,zheng2019frap,oroojlooy2020attendlight,wei2018intellilight,wei2019presslight,d2021timeloss} through the advanced reinforcement learning (RL) methods~\cite{mnih2015human,schulman2017proximal,liang2021integrated}. A proper reward function is required for applying RL to the TSC problem and improving the policy. Since the travel time is only feasible after a vehicle leaves the road network and fails to provide instant/dense feedback to the TSC policy in time during its journey, many previous works~\cite{varaiya2013maxpressure,wei2019presslight,wei2019colight,xu2021hierarchically} draw the traditional characteristics of traffic intersections for the reward design, such as traffic pressure and queue length at the intersection. However, most of these heuristic rewards may be biased from the ultimate goal of TSC, i.e., the average travel time minimization. The introduced biases could lead the RL methods to adjust the TSC policy in an incorrect direction.

Apart from the reward design, it is also critical to endow the RL agents with the capability of precisely predicting the future dynamics of the environment to make well-founded decisions~\cite{van2016deep,fujimoto2018addressing,yang2016optimized,lou2020probabilistic}. However, it is non-trivial for an RL agent to capture the future traffic dynamics of the intersections in the context of TSC~\cite{chen2022bidirectional}. The future arriving vehicles of one intersection may be running in a \textbf{distant} intersection at present. To this end, the \textbf{local} observations of either the intersection itself (i.e., a snapshot of vehicles at the intersection)~\cite{varaiya2013maxpressure,wei2019presslight,d2021timeloss} or neighboring intersections~\cite{wei2019colight,xu2021hierarchically} might be insufficient to predict the long-horizontal traffic dynamics of the intersection. Additionally, the location and traffic flow trend of an intersection also play a role in predicting the traffic dynamics of the intersection. For example, downtown intersections tend to be more crowded than suburban intersections. For example, a large number of vehicles entering an empty intersection may indicate the beginning of the rush hour. Therefore, the non-local observations and the location of the intersections, and historical traffic conditions are all important for a TSC agent to estimate future traffic situations more accurately.

To address the issues discussed above, we propose a novel RL-based TSC method named \textit{DenseLight}, which improves traffic light coordination by exploiting dense information from both an unbiased reward and non-local intersection information fusion. Specifically, to provide dense and unbiased feedback for the policy improvement, we propose an equivalent substitute for travel time, i.e., the gap between the ideal traveling distance (i.e., the ideal distance a vehicle could have traveled at the full speed during its journey) and the actual traveling distance during the whole journey, namely Ideal-Factual Distance Gap (IFDG). Since the length of the factual journey of a vehicle is fixed according to its traveling lanes of the road network, minimizing IFDG is equivalent to minimizing travel time. Most importantly, IFDG can be calculated at each intersection and at each timestep. Therefore, IFDG can also provide instant feedback on the effectiveness of the control policy at each intersection.

Besides an unbiased and dense reward, DenseLight also features a \textit{Non-local enhanced Traffic Signal Control}~(NL-TSC) agent to benefit TSC from the spatial-temporal augmented observation and the non-local fusion network architecture. Specifically, the NL-TSC agent supplements the original observation of an intersection with its location information and the previous observation so that each intersection can customize its own signal plans w.r.t. historical and spatial information. As for facilitating the TSC agents with a better awareness of the future traffic dynamics affected by other intersections, a non-local branch is proposed to enhance the local features of each intersection with the features of the non-local intersections. By learning to communicate the non-local information across non-local intersections, the NL-TSC agent can better predict the long horizontal traffic condition of each intersection and can thus make better coordination at present.

Overall, our contributions are three-fold: {\bf{1)}} we propose a novel RL-based TSC method, i.e., DenseLight, which is optimized by an unbiased and dense reward termed IFDG; {\bf{2)}} to better model the future accumulated IFDG of each intersection, we develop the NL-TSC agent, effectively gathering spatial-temporal features of each intersection and propagating the non-local intersection information to improve the multi-agent RL policy; {\bf{3)}} comprehensive experiments conducted on different real-world road networks and various traffic flows show that DenseLight can consistently outperform traditional and advanced RL-based baselines.

\section{Related Works}

\paragraph{{Conventional traffic signal control methods.}} Traditional traffic signal control methods~\cite{little1981maxband,roess2004traffic,koonce2008traffic} typically set traffic signal plans with fixed cycle lengths, phase sequences, and phase splits. They heavily relied on expert knowledge and hand-crafted rules. Adaptive traffic signal control methods formulated the task as an optimization problem and made decisions according to the pre-defined signal plans and real-time data~\cite{hunt1981scoot,luk1982scats,mirchandani2001real,cools2013self,hong2022traffic}. Recently, researchers have made progress by comparing the requests from each phase through different traffic representations~\cite{varaiya2013maxpressure,zhang2021expression,zhu2022hybrid,liu2023cross}.

\paragraph{{RL-based traffic signal control methods.}} 
For individual intersection signal control, some studies investigated RL environmental settings~\cite{van2016coordinated,wei2018intellilight,wei2019presslight,zhang2021expression}, agent network design~\cite{zheng2019frap}, and policy transfer learning~\cite{zang2020metalight,liu2018hierarchically,oroojlooy2020attendlight,liu2019deep,liu2021semantics} to optimize the average travel time of all vehicles. Since the travel time can not be obtained until a vehicle leaves the road network, these methods presumed some objectives to be equivalent to maximizing the traffic throughput, thus minimizing travel time. For example, ~\cite{van2016coordinated} utilized a combination of vehicle delay and waiting time as the reward, and PressLight~\cite{wei2019presslight} introduced the pressure of intersection to be a new objective. However, these optimization objectives are not always aligned with the minimization of the average travel time. To resolve this conflict, ~\cite{d2021timeloss} defined an instant time loss for vehicles to guide the next action selection. This reward is the average loss of time at the moment when selecting traffic signals. Therefore, time loss is not able to reflect the later change in traffic congestion caused by the newly selected signals. Differently, our IFDG integrates speed over time after selecting a signal, and thus IFDG can reflect the effect of the signal.

\paragraph{{Traffic signal coordination.}}To coordinate multiple intersections, a centralized controller could select actions for all intersections ~\cite{prashanth2010reinforcement,tan2019cooperative,van2016coordinated} but may suffer from the curse of dimension in combinatorially large action space. Therefore, some studies modeled single agents in a decentralized way and scaled up to multi-intersection settings by parameter sharing~\cite{chen2020mplight}. However, ignoring the influence of other intersections may affect the objective optimization above the whole network. Some studies used centralized training and decentralized execution~\cite{pmlr-v80-rashid18-QMIX,pmlr-v97-son19-QTRAN,wang2020large,tan2019cooperative}; 
some gathered neighboring information by concatenating neighborhood observations~\cite{liu2018global,chu2019multi,liu2022causal} or communicating through graph neural network~\cite{wei2019colight,Guo2021MaCAR,zeng2021graphlight,huang2021network,liu2022tcgl}.
However, these methods either treated other intersections equally or only considered the neighborhood, ignoring useful information from distant intersections.

%

\section{Preliminaries}
\label{preliminary}
\subsection{Traffic Signal Control} \label{sec:pre_tsc}
Conventionally, a TSC task $\mathcal{D}$ includes a road network and a set of vehicle flows. A road network is usually composed of $N$ multiple intersections $I=[1, ..., N]$ which can regulate the vehicles in their entering lanes via selecting different traffic signals. 
Specifically, an \textbf{entering lane} of an intersection is a lane where the vehicles enter the intersection from the north, south, west, or east direction. We use $L_{i}^\text{in}$ to represent all the entering lanes of the $i$-th intersection. And the \textbf{exiting lanes} $L_{i}^\text{out}$ are the downstream lanes of the $i$-th intersection where the vehicles leave the intersection. A \textbf{traffic movement} is the traffic moving from an entering lane $l_{i,j}^\text{in} \in L_{i}^\text{in}$ to an exiting lane $l_{i,k}^\text{out} \in L_{i}^\text{out}$, which is generally categorized as a left turn, through and right turn.
We define a combination of non-conflicting traffic movements as a \textbf{phase} $a$. An intersection with phase $a$ gives the corresponding traffic movements the priority to pass through under a green light while prohibiting the others. According to the rules in most countries, right-turn traffic is allowed to pass unrestricted by signals. By convention~\cite{zhang2021expression}, we only consider a 4-way intersection, resulting in 12 movements and 4 candidate phases.

The goal of TSC is to learn a control policy $\pi_\theta$ parameterized by $\theta$ for all intersections in the road network to minimize the average travel time of all the vehicles ${X}$ in a road network, within a finite period $T_\text{TSC}$, i.e. $\min_\theta \frac{1}{|{X}|} \sum_{x \in {X}} x_l - x_e,$
where $x_e$ and $x_l$ are the moments the vehicle $x$ enters and leaves the road network, respectively. Following the conventional setting in \cite{zheng2019frap,zhang2021expression}, each phase persists for a fixed phase duration $T_\text{phase}$. To avoid confusion, we use $d$ to denote the $d$-th decision, so that $t_{d+1} - t_d = T_\text{phase}$. We use $D$ to denote the total number of decision-making steps in each TSC episode, where $T_\text{TSC}$ = $D \times T_\text{phase}$. We use $v_t(x)$ to stand for the speed of vehicle $x$ at time $t$, and $v_{\rm max}$ to represent maximum speed.


\subsection{Reinforcement Learning for TSC}
In this paper, we consider a standard RL framework where an agent selects the phase for each intersection after every $T_\text{phase}$ seconds. At $t_d$, a TSC agent observes the intersection state $s_d$ from an intersection. Then, the agent chooses a phase $a_d$ according to the policy $\pi$. The environment returns the reward $r_d$ and the next observation $s_{d+1} \sim \mathcal{T}(s_d, a_d)$, where $\mathcal{T}$ is the environment transition mapping. The return of a trajectory after the $d$-th phase is $\eta_{d}=\sum_{k=0}^\infty \gamma^k r_{d+k}$,
where $\gamma \in [0, 1)$ is the discount factor. RL aims to optimize the policy to maximize the expected returns of future trajectories after observing $s_d$.

\paragraph{Elements in observation and reward of TSC.} In most previous TSC methods, observations and rewards originated from the traffic statistic characteristics of intersections, including the following examples. \textbf{1) Pressure}: the difference between the number of upstream and downstream waiting vehicles (with a speed less than 0.1m/s) of the intersection in a time step; \textbf{2) Queue length}: the average number of the waiting vehicles in the incoming lanes in a time step; \textbf{3) Time loss}: sum of the time delay of each vehicle, i.e., $1-v_{t_d}/v_\text{max}$, at the moment of phase selection $t_d$; \textbf{4) Step-wise travel time}: the local travel time at each intersection within the duration $T_{\rm phase}$. As proved in Appendix, step-wise travel time is an unbiased reward for TSC, however, provides sparse feedback for each intersection at each time step.

\begin{figure}[t]
    \centering
    \includegraphics[clip=True, trim={0 50 565 0}, width=\linewidth]{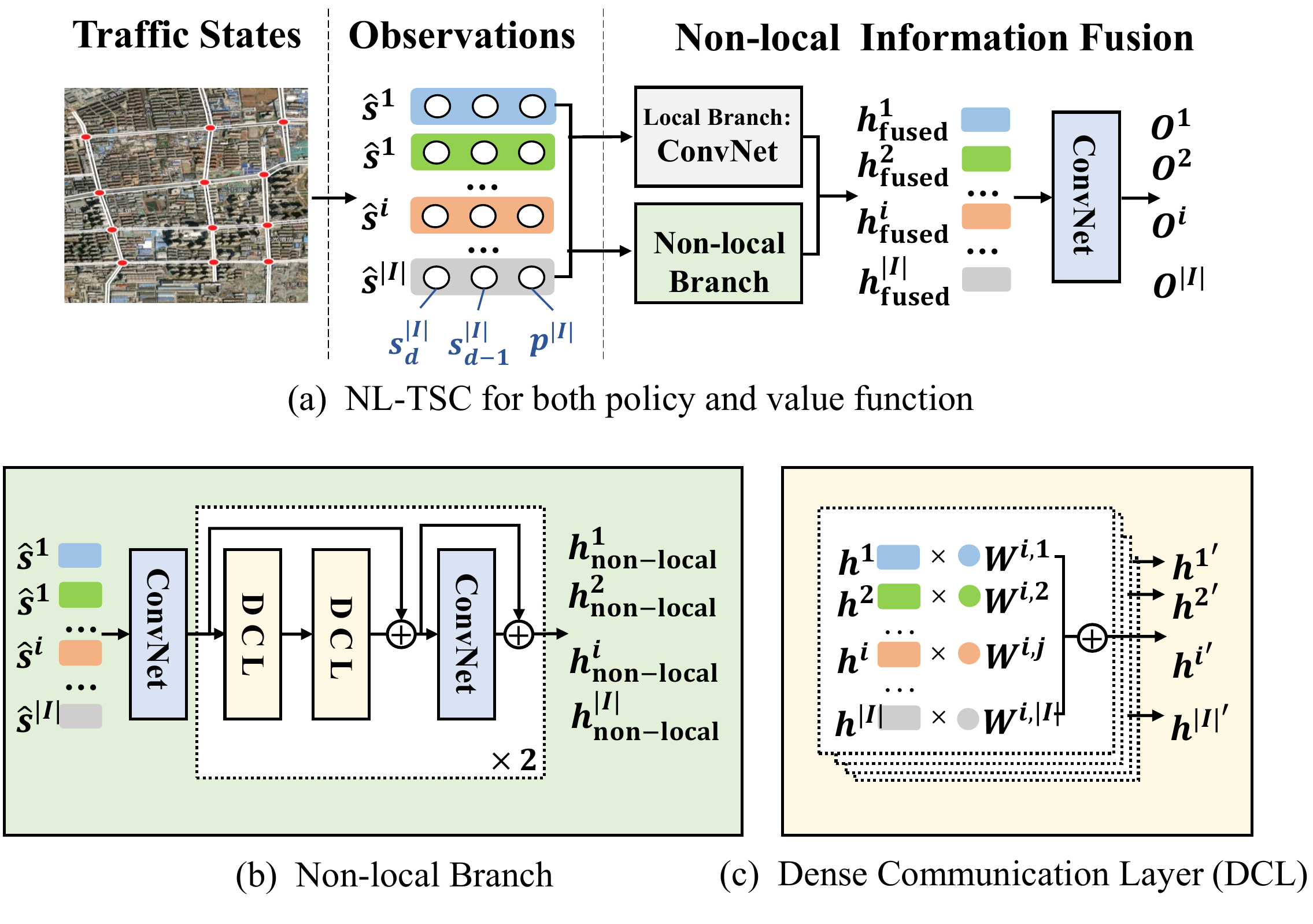}
    \caption{The policy/value estimator architecture of NL-TSC agent. Based on the observations from the road network, an NL-TSC agent predicts action distribution/value through a local branch and a non-local branch to realize non-local information fusion. The local branch adopts a ConvNet that extracts features of each intersection using its own observation alone; the non-local branch consists of Dense Communication Layers (DCL) and ConvNets, integrating non-local features into the representation of each intersection.}
    \label{fig:network_structure}
\end{figure}

\section{DenseLight}
\label{methods}

In our paper, we propose a novel RL-based TSC method, namely, DenseLight, which contains two key ingredients, i.e., \textit{Ideal-Factual Distance Gap} reward and \textit{Non-local enhanced Traffic Signal Control} agent, as shown in Fig. \ref{fig:network_structure}. In this section, we elaborate on each of them in detail.

\subsection{Ideal-Factual Distance Gap} \label{method_IFDG}
As mentioned in Sect.~\ref{preliminary}, the goal of TSC is to search for a traffic signal control policy to minimize the average travel time of vehicles. Before RL takes place in TSC, traditional methods, such as MaxPressure~\cite{varaiya2013maxpressure}, have been developed and widely adopted in real-world road networks. The elements of traditional TSC methods like pressure and queue length have inspired the later development of RL-based methods~\cite{wei2019presslight,wei2019colight,chen2020mplight}. 
By trivially adopting traditional elements as a negative reward, prior RL-based methods have witnessed a great improvement upon the traditional methods by a significant margin. However, these heuristic rewards are usually designed empirically and not derived from the target. Optimizing with these rewards might hamper the RL agent from improving in the correct direction.

To develop an unbiased and dense reward function for TSC, we propose the \textit{Ideal-Factual Distance Gap} (IFDG) reward function. As the name suggests, IFDG characterizes the gap between the ideal distance that the vehicles could have traveled at full speed and the factual distance affected by the phase decision of the RL agent. Ideally, if vehicles are running at full speed, IFDG will be zero. We use the negative IFDG as the reward $r_{\text{IFDG}, d}^i$ for $i$-th intersection received after $d$-th decision, and $x^i \in X^i$ to represent each vehicle at $i$-th intersection. Formally, 
\begin{align}
    &r_{\text{IFDG}, d}^i = -\sum_{\forall x^i: t_d < x_l^i \leq t_{d+1}} \int_{\max \{t_d, x_e^i\}}^{x_l^i} v_{\rm max} - v_t(x^i) {\rm d} t \label{equ:IFDG} \\
    &-\sum_{\forall x^i: (t_{d+1} < x_l^i) \wedge (x_e^i \leq t_{d+1})} \int_{\max \{t_d,x_e^i\}}^{t_{d+1}} v_{\rm max} - v_t(x^i) {\rm d} t. \nonumber
\end{align}
According to the Eq.~(\ref{equ:IFDG}), the vehicles that either stay or leave the intersection will contribute their distance gaps to the reward. And since the distance gap is sensitive to traffic congestion, $r_{\text{IFDG},d}^i$ can effectively reflect the difference in different degrees of traffic congestion caused by different phases. 
More importantly, IFDG turns out to be an equivalent substitute for travel time:
\begin{align}
    \sum_{d=0}^{D-1} \sum_{i \in I} &r_{\text{IFDG}, d}^i = -\sum_{x \in X} \int_{x_e}^{x_l} (v_\text{max} - v_t(x)) {\rm d} t \nonumber \\
    &= \sum_{x \in X}\int_{x_e}^{x_l} v_t(x) {\rm d} t -\sum_{x \in X} v_\text{max} \cdot (x_l - x_e) \nonumber \\
    &= \text{constant} - v_\text{max} \sum_{x \in X} (x_l - x_e), \label{equ:factualpath}
\end{align}
where the first equation is the sum of the IFDG of all vehicles. The $\sum_{x \in X}\int_{t=x_e}^{x_l} v_t(x) {\rm d} t$ in Eq.~(\ref{equ:factualpath}) is the sum of the traveling distance of all vehicles, which is a constant since each vehicle has a fixed traveled path in our task. And the right-hand side of Eq.(\ref{equ:factualpath}) is linear correlative with the travel time. Thus, our IFDG provides dense feedback for RL improvement and guides the improvement in an unbiased direction.

\subsection{Non-local Enhanced TSC Agent}
Considering the Markov property in RL, the decision-making and the value estimation of an agent require the observation of RL policy to be self-contained. Due to the complex relations between intersections of TSC, it is not sufficient to take the local observation of the intersection to estimate the policy's future value.
In this paper, we propose an \textit{Non-local enhanced Traffic Signal Control} (NL-TSC) agent, which takes advantage of spatial-temporal augmented observation and non-local traffic information fusion to make a better decision and estimate a more accurate value for each intersection.

\paragraph{{Spatial-temporal augmented observation.}} In a road network, the dynamics of traffic congestion at an intersection generally vary across different areas and different periods. For example, the traffic congestion in downtown intersections is usually heavier than that in other areas. And the traffic tends to be more crowded at the beginning of the rush hour and more smooth at the end of it. Therefore, the TSC agent requires not only the current traffic conditions observed in each intersection but also the location and the tendency of the traffic congestion of an intersection to make more comprehensive decisions. To this end, we propose a spatial-temporal augmented observation for $i$-th intersection by concatenating the original observation $s_d^i$ at time $t$ with its position encoding $p^i$ and its previous observation $s_{d-1}^i$:
\begin{align}
  \hat{s}_d^i = \text{concatenate}\{s_{d}^i, s_{d-1}^i, p^i\}, \label{equ:st} 
\end{align}
where $\text{concatenate}\{...\}$ puts all vectors in the bracket into one vector and $p^i$ is a 2-D position encoding implemented as ~\cite{dosovitskiy2020image}. 
Including the position information allows NL-TSC agents of each intersection to make decisions according to their locations. Moreover, augmented with the previous observation, the changes in traffic conditions can be captured to better predict future traffic dynamics.

\paragraph{{Non-local information fusion.}} On the TSC task, vehicles arriving at one intersection may come from either nearby or distant intersections. In this sense, the relations between intersections are complex and entangled. To provide sufficient information to estimate long-horizontal values, we propose a non-local branch that can automatically determine what information to communicate among non-local intersections. Specifically, the NL-TSC agent first extracts features for each intersection $i$ by \textbf{a stack of multiple $1\times 1$ convolutional layers} (abbr. ConvNet) (Eq.~(\ref{equ:embed})). After that, the non-local branch propagates the features of non-local-intersections by a \textit{Dense Communication Layers} (DCL) which is parameterized by $W\in \mathbb{R}^{|I|\times |I|}$, as shown in Fig.~\ref{fig:network_structure}(c). The features of non-local intersections are fused and integrated into the original representations of each intersection according to Eq.~(\ref{equ:nonlocal}). Then, the final non-local features ${h^i}_\text{non-local}$ for each intersection are processed by another ConvNet, as formulated in Eq.~(\ref{equ:process}).
\begin{align}
    &h^i = \text{ConvNet}_\text{embed}(\hat{s}^i). \label{equ:embed} \\
    &{h^i}' = h^i + W^i \times [h^1, h^2, ..., h^{|I|}]. \label{equ:nonlocal} \\
    &{h^i}_\text{non-local} = {h^i}' + \text{ConvNet}_\text{process}({h^i}'). \label{equ:process}
\end{align}
However, as the size of the road network increases, $W$ tends to be over-parameterized and degrades the expression ability of the extracted non-local features. To address this problem, we use two consecutive DCLs with two smaller parameters $W_a\in \mathbb{R}^{|I|\times M}$ and $W_b\in \mathbb{R}^{M\times |I|}$ to replace the original DCL with $W$, so that $W = W_a \times W_b$, as demonstrated in Fig.~\ref{fig:network_structure}(b). The above operations (Eq.~(\ref{equ:nonlocal}) and ~(\ref{equ:process})) are repeated twice to better fuse the non-local information. 
To allow the model to automatically fuse local and non-local information, the NL-TSC agent also extracts the local features focusing on the intersection itself:
\begin{align}
    h^{i}_\text{local} = \text{ConvNet}_\text{local}(\hat{s}^i).
\end{align}
Finally, both local and non-local features are forwarded to the final ConvNet to predict the categorical action distribution or the future policy value for each intersection:
\begin{align}
    h^{i}_\text{fused} &= \text{concatenate}\{h^{i}_\text{non-local}, h^{i}_\text{local}\}, \\
    o^i &= \text{ConvNet}_\text{final}(h^{i}_\text{fused}).
\end{align}
$o^i$ can be either the action logits $\log \pi_\theta(a|\hat{s})$ or the value $V_\phi(\hat{s})$ where $\theta$ and $\phi$ are the networks parameters of the policy and value estimator, respectively. 
The structure of either policy or value estimator of the NL-TSC agent is sketched in Fig.~\ref{fig:network_structure}(a). As for RL-optimization, we adopt the well-known and stable policy gradient algorithm, proximal policy optimization (PPO-Clip)~\cite{schulman2017proximal}.

\section{Experiments}
\subsection{Experiment Setting}
The experimental environment is simulated in CityFlow \cite{zhang2019cityflow}, which simulates vehicles behaviors every second, provides massive observations of the road network for the agents, and executes the TSC decisions from the agents. Following the existing studies~\cite{wei2019presslight,zhang2021expression}, each episode is a 3600-second simulation (i.e., $T_\text{TSC}=3600$), and the action interval $T_\text{phase}$ is 15 seconds, then the number of decision-making steps, i.e., $D$, is 3600/15 = 240. By convention, a three-second yellow signal is set when switching from a green signal to a red signal.

\paragraph{{Datasets.}} Experiments use four real-world road networks. \textbf{i) Jinan12}: with $4\times3$ (4 rows, 3 columns) intersections in Jinan, China; \textbf{ii) Hangzhou16}: with $4\times4$ intersections in Hangzhou, China; \textbf{iii) NewYork48}: with $16\times3$ intersections in New York, USA; \textbf{iv) NewYork196}: with $28\times7$ intersections in New York, USA. For each road network, the corresponding traffic flow data are collected and processed from multi-sources with detailed statistics recorded in Tab.~\ref{tab:datasets}. To evaluate the performance of the TSC methods on various traffic conditions, we synthesize a novel dataset $\mathcal{D}_\text{JN12(f)}$ with high average traffic volumes and high variances (which indicates the fluctuating traffic flow) by randomly re-sampling vehicles from the real traffic flow data and adding them back to the original data with re-assigned enter time.

\begin{table}[t]
    \setlength{\tabcolsep}{3pt}
  \centering
\footnotesize
    \begin{tabular}[b]{llllll}
    \toprule
    \multirow{2}{*}{Road Network} & \multirow{2}{*}{Traffic Flow} & \multicolumn{4}{c}{Arrival Rate (vehicles/300s)}\\
    \cmidrule(r){3-6}
    & & Mean & Std. & Max & Min\\
    \midrule
    \multirow{2}{*}{Jinan12}
    & $\mathcal{D}_\text{JN12}$
    & 526.64 & 86.70 & 676 & 256  \\
    & $\mathcal{D}_\text{JN12(2)}$
    & 363.61 & 64.10 & 493 & 233  \\
    & $\mathcal{D}_\text{JN12(f)}*$
    & 639.11 & 177.63 &	1036  &	281   \\
    \midrule
    \multirow{2}{*}{Hangzhou16}
    & $\mathcal{D}_\text{HZ16}$
    & 250.71& 	38.20 & 	335 & 	208  \\
    & $\mathcal{D}_\text{HZ16(f)}$
    & 572.48&	293.39&	1145&	202  \\
    \midrule
    NewYork48
    & $\mathcal{D}_\text{NY48}$
    & 236.42 &	8.13  &	257	  & 216  \\
    \midrule
    NewYork196
    & $\mathcal{D}_\text{NY196}$
    & 909.90 &	71.22  & 	1013 &	522 \\
    \bottomrule
    \end{tabular} 
    \caption{Data statistics of the traffic flow datasets. Dataset with the symbol * is generated based on the real data with an increasing volume of vehicles, and the others are open real-world datasets.}
  \label{tab:datasets}
\end{table}

\begin{table*}[t]
  \centering
    \begin{tabular}{cccccccc}
    \toprule
    Method & $\mathcal{D}_\text{JN12}$ & $\mathcal{D}_\text{JN12(2)}$ & $\mathcal{D}_\text{JN12(f)}$  & $\mathcal{D}_\text{HZ16}$ & $\mathcal{D}_\text{HZ16(f)}$ & $\mathcal{D}_\text{NY48}$ & $\mathcal{D}_\text{NY196}$\\
    \midrule
    Fixed-Time & 415.63 & 351.84 & 413.24 & 475.44 & 393.92 & 1070.45 &  1506.85 \\
    MaxPressure & 259.87 & 229.74 & 279.88 & 268.37 & 336.78 & 775.63 &   1178.71 \\
    Efficient-MP & 256.00 & 224.45 & 277.20 & 264.25 & 315.38 & 293.27 &   1122.32 \\
    Advanced-MP & 241.43 & 225.71 & 262.32 & 263.49 & 309.29 & 197.84 &   1060.36 \\ \midrule
    FRAP
    & 273.50 & 249.35 & 298.99 & 287.07 & 352.20 
    & 180.04 & 1241.54\\
    MPLight
    & 278.27 & 251.39 & 303.36 & 297.80 & 349.29
    & 1841.86 & 1909.92 \\
    CoLight
    & 252.87 & 235.09 & 278.59 & 276.08 & 329.56
    & 168.12  & 988.78 \\
    Advanced-MPLight
    & 239.10 & 219.21 & 250.36 & 255.06 & 306.25
    & 1617.80 & 1339.09 \\
    Advanced-CoLight
    & 232.16 & 217.20 & 251.21 & 251.44 & 303.15
    & 160.25 & 1004.52 \\
    TimeLoss-FRAP
    & 234.34 & 221.90 & 249.62 & 258.47 & 321.06
    & 896.25 & 1143.90 \\ \midrule
    DenseLight (Ours)
   & \textbf{226.97} & \textbf{215.82} & \textbf{239.58}
    & \textbf{248.43} & \textbf{272.27} 
    & \textbf{156.30} & \textbf{803.42} \\
    \bottomrule
    \end{tabular}
 \caption{The average travel time results of baselines across different traffic flow datasets in different road networks.}
  \label{tab:comparing}
\end{table*}

\paragraph{{Baselines.}} To evaluate the effectiveness of our method, we compare our DenseLight with the traditional and RL-based approaches. The results of the baselines are obtained by re-running their corrected open codes. \textit{The traditional methods} are described in the following. \textbf{Fixed-Time~\cite{koonce2008traffic}:} Fixed-Time control consists of pre-defined phase plans that are fixed in duration; \textbf{MaxPressure~\cite{varaiya2013maxpressure}:} an adaptive policy that greedily selects the phase with the maximum pressure; \textbf{Efficient-MP~\cite{wu2021efficient}:} an adaptive policy that selects the phase with the maximum efficient pressure. Efficient pressure is the difference of average queue length between the upstream and downstream of each traffic movement; \textbf{Advanced-MP~\cite{zhang2021expression}:} an adaptive method that defines the number of running vehicles within an effective range near the intersection as the request of the current phase, and defines the pressure as the requests of other phases. The policy selects the phase with the maximum request. 
And the \textit{advanced RL-based methods} are as follows. \textbf{FRAP~\cite{zheng2019frap}:} FRAP designs a network for traffic signal control that models the competition relations between different phases based on the demand prediction for each phase with queue length as the reward; \textbf{MPLight~\cite{chen2020mplight}:}  MPLight integrates pressure into the observation and uses the reward and agent as FRAP~\cite{zheng2019frap}; \textbf{CoLight~\cite{wei2019colight}:} CoLight uses a graph attention network to learn the influences from neighboring intersections and adopts the length of vehicles in entering lanes as the reward; \textbf{Advanced-MPLight~\cite{zhang2021expression}:} Based on MPLight, Advanced-MPLight uses the current phase and the advanced traffic states (including efficient pressure and the number of vehicles within an effective range) as an intersection observation; \textbf{Advanced-CoLight~\cite{zhang2021expression}:} Advanced-CoLight adds the advanced traffic states to the observation of CoLight; \textbf{TimeLoss-FRAP~\cite{d2021timeloss}:} TimeLoss-FRAP uses the FRAP~\cite{zheng2019frap} agent with the instant time loss of vehicles in the entering lanes as the observation and reward; \textbf{DenseLight (Ours):} The component of observation $s_d^i$ at the intersection $i$ and step $d$ is the same as Advanced-MPLight. DenseLight is optimized with IFDG reward and NL-TSC agent.

\paragraph{{Training details.}} As for DenseLight, the training hyper-parameters follow the defaults in PPO-Clip~\cite{schulman2017proximal} implemented in the open source platform~\cite{weng2021tianshou}. In brief, the number of episodes in $\mathcal{B}_k$ for $k \in [0, 500)$ is 2. The size of all hidden layers of neural networks is 64. Both the local branch of NL-TSC and $\text{ConvNet}_\text{process}$ are ConvNets with 2 $1\times 1$ convolutional layers. Other ConvNets in NL-TSC are single $1\times 1$ convolutional layers. The size of the training batch is 64. The learning rate is 3e-4 at the beginning and linearly decays to 0. $M$ of $W_a$ and $W_b$ is set as $|I|$. For $\mathcal{D}_\text{NY196}$, the batch size is set as 16 to reduce the memory occupancy, $M$ is set as $0.1\times|I|$ to prevent over-parameterization, and the $28\times7$ road network is divided into four $7\times7$ road networks to reduce the dimension of the state. These measures are adopted to ease the optimization without causing conflicts with our contributions. The dimension of position encoding is 16. Following the conventional settings in ~\cite{chen2020mplight,zhang2021expression}, all results in the tables of experiments are the average of the final 10 evaluation episodes.

\subsection{Results}
We test each method in 60-minute simulations separately and report the average performance of the last ten episodes. 
We have following findings from the comparison results in Tab.~\ref{tab:comparing}.

\paragraph{{Consistent improvement.}} As shown in Tab.~\ref{tab:comparing}, our DenseLight achieves consistent improvement by a clear margin across different road networks and traffic flows. Particularly, DenseLight hits a travel time of 800 seconds on $\mathcal{D}_\text{NY196}$ while the previous state-of-the-art result is about 1000 seconds, saving about 20\% of the time. Another significant improvement can be observed under $\mathcal{D}_\text{HZ16(f)}$ where DenseLight achieves nearly 10\% improvement over the previous best result.


\begin{figure}[!t]
\centering
\subcaptionbox{IFDG reward \label{fig:ifdg_reward}}
{\includegraphics[width=0.49\linewidth]{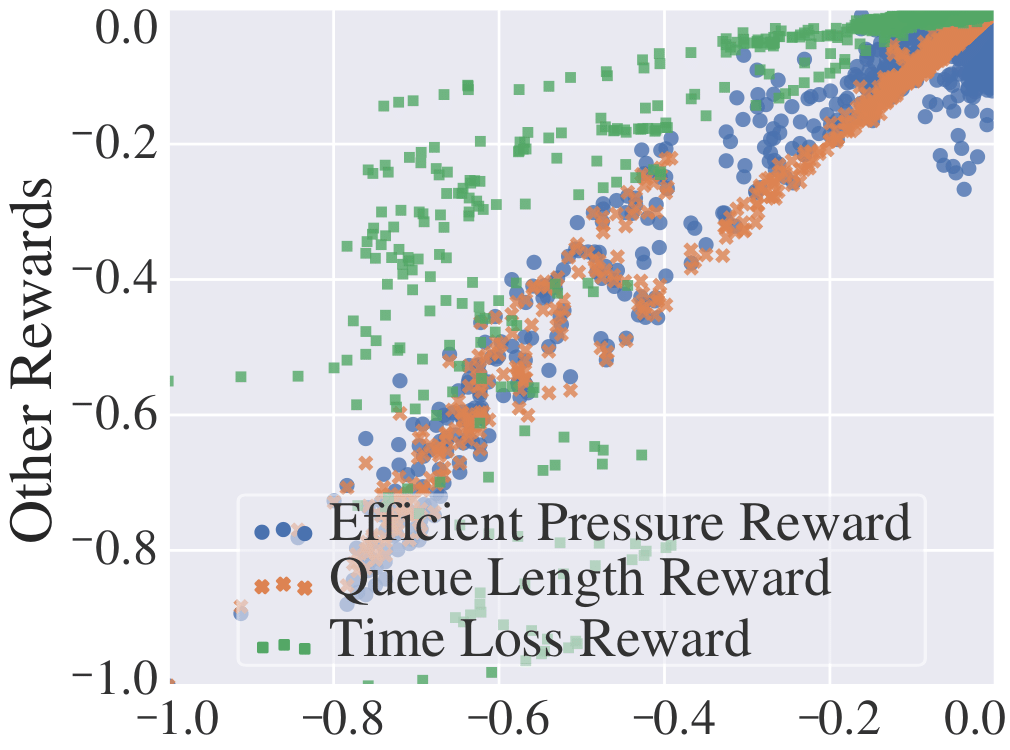}} 
\subcaptionbox{Step-wise travel time \label{fig:stt_reward}}
{\includegraphics[width=0.49\linewidth]{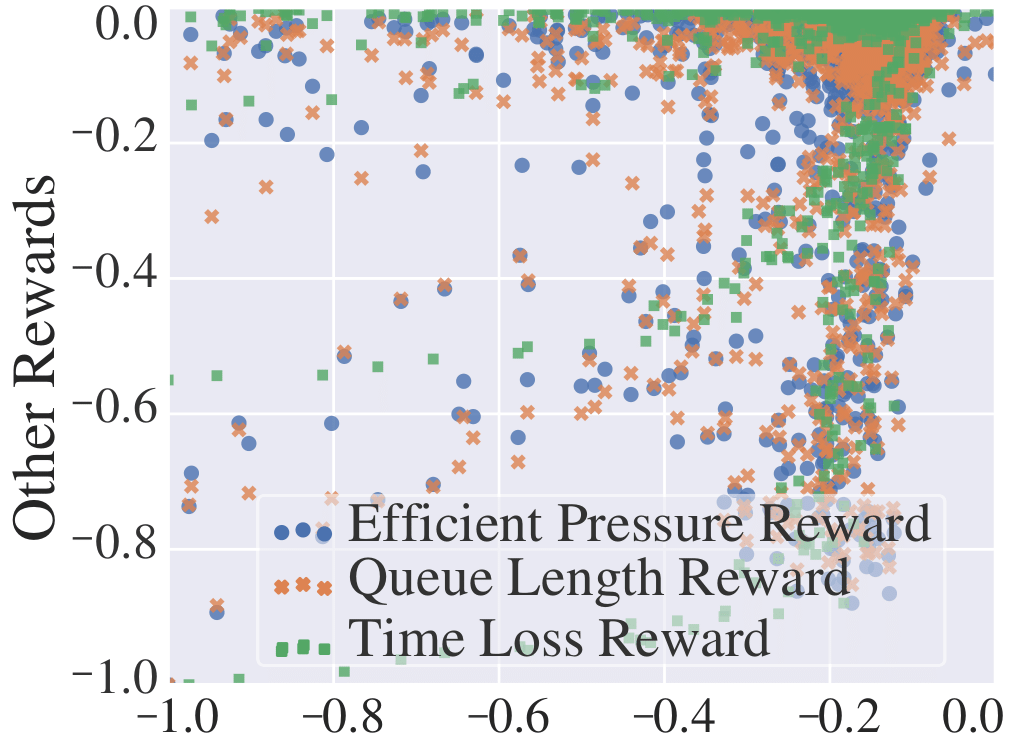}}
\caption{Relations between the normalized IFDG reward/step-wise travel time reward and other normalized rewards.}
\label{fig:visualization}
\end{figure}

\paragraph{{Generalization to various traffic flows.}} We observe that under two existing flow data $\mathcal{D}_\text{JN12}$ and $\mathcal{D}_\text{JN12(2)}$ with the Jinan12 road network, different methods have close performances. However, on another similar road network, i.e., Hangzhou16, DenseLight can obtain a more significant improvement under $\mathcal{D}_\text{HZ16(f)}$. The potential reason is that the traffic flow in  $\mathcal{D}_\text{HZ16(f)}$ shows a more diverse pattern than that of the Jinan12 road network. As shown in Tab.~\ref{tab:datasets}, 
the numbers of arriving vehicles in every 300 seconds in $\mathcal{D}_\text{HZ16(f)}$ vary considerably.
To further verify this point, we have designed a fluctuating flow $\mathcal{D}_\text{JN12(f)}$ with its detailed statistics presented in Tab.~\ref{tab:datasets}. From the results of $\mathcal{D}_\text{JN12(f)}$, we observe that the improvement of DenseLight becomes more evident.

\begin{table*}[t]
\setlength{\tabcolsep}{1.5mm}
  \centering
   {
    \begin{tabular}{ccccccccc}
    \toprule
    Component & Method & $\mathcal{D}_\text{JN12}$ & $\mathcal{D}_\text{JN12(2)}$ & $\mathcal{D}_\text{JN12(f)}$ & $\mathcal{D}_\text{HZ16}$ & $\mathcal{D}_\text{HZ16(f)}$ & $\mathcal{D}_\text{NY48}$ & $\mathcal{D}_\text{NY196}$\\
    \midrule
    \multirow{5}{*}{Reward} & EfficientPressure~\cite{zhang2021expression}
    & 230.31 & 217.30 & 244.50
    & 251.03 & 334.55
    & 1556.77 & 1106.46 \\
    &Queue Length~\cite{wei2019colight}
    & 229.76 & 217.30 & 247.25
    & 250.53 & 293.42 
    & 162.43 & 1026.24 \\
    &TimeLoss~\cite{d2021timeloss}
    & 233.88 & 217.07 & 257.78
    & 250.49 & 313.18 
    & 161.67 & 1049.19 \\
    &Step-wise Travel Time 
    & 230.10 & 217.78 & 244.47
    & 253.16 & 288.41
    & 162.91 & 1056.36 \\
    &Ideal-Factual Distance Gap
    & 229.60 & 216.77 & 243.50
    & 250.38 & 276.75 
    & 161.02 & 963.21 \\ 
    \midrule
    \multirow{3}{*}{S.T. Aug.} &Consecutive Observation (C)
    & 229.04 & 217.08 & 242.71
    & 250.30 & 274.53
    & 160.90 & 1052.93 \\
    &Position Encoding (P)
    & 228.73 & 216.70 & 242.29
    & 249.61 & 276.48
    & 159.59 & 954.99 \\
    &Spatial-temporal Augmentation (C+P)
    & 228.20 & 216.31 & 240.03
    & 249.38 & 274.16
    & 159.50 & 953.48 \\ \midrule

    \multirow{3}{*}{Non-local}&1-hop
    & 228.17 & 216.12 & 240.21 & 249.17 & 288.81 & 157.78 & 926.85 \\
    &2-hop
    & 228.23 & 216.31 & 240.36 & 249.10 & 284.52 & 158.18 &  931.62 \\
    &Transformer~\cite{vaswani2017attention}
    & 227.79 & 216.12 & 240.44
    & 249.14 & 274.04 & 158.66 & 917.25 \\ 
    \midrule
     &DenseLight
    & \textbf{226.97} & \textbf{215.82} & \textbf{239.58}
    & \textbf{248.43} & \textbf{272.27} 
    & \textbf{156.30} & \textbf{803.42} \\
    
    \bottomrule
    \end{tabular} 
 \caption{The average travel time results of ablation studies, ranging from rewards to agent components. For briefness, we use S.T. Aug. to stand for spatial-temporal augmentation.}
  \label{tab:ablation}
  }
\end{table*}
\subsection{Ablations and Analyses}

In this part, we sequentially add modules for ablation. 

\begin{figure}[!t]
\centering
\includegraphics[width=0.95\linewidth]{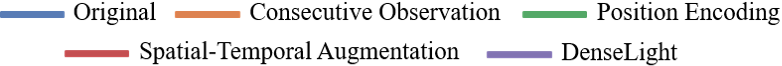} \\
\subcaptionbox{$\mathcal{D}_\text{HZ16(f)}$ \label{fig:hangzhou2_valueloss}}
{\includegraphics[width=0.49\linewidth]{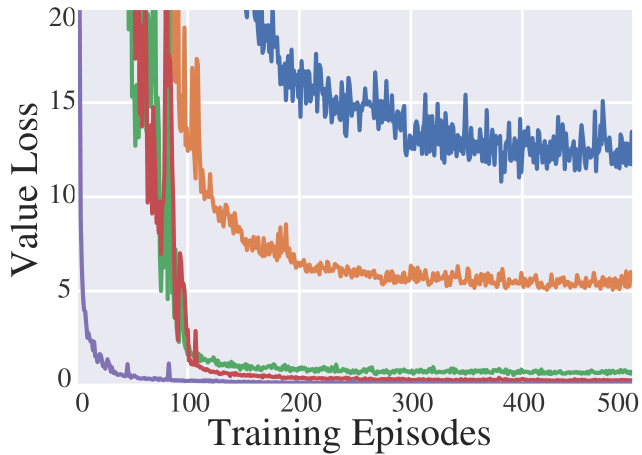}}
\subcaptionbox{$\mathcal{D}_\text{NY195}$ \label{fig:newyork_valueloss}}
{\includegraphics[width=0.49\linewidth]{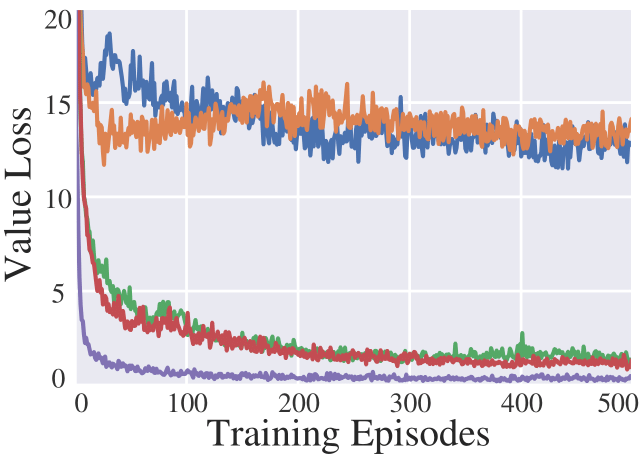}}
\caption{Errors of the value prediction with different components.}
\label{fig:valueloss}
\end{figure}

\subsubsection{{Rewards}}
To evaluate the effect of reward, we compare our IFDG reward with other rewards. In the experiments, all the elements except for the reward are strictly controlled to be the same. The upper part of Tab.~\ref{tab:ablation} shows that our IFDG obtains the best performance. As the size of the road network and the variance of flows increase, the advantage of IFDG becomes more significant. Especially on $\mathcal{D}_\text{HZ16(f)}$ and $\mathcal{D}_\text{NY196}$, our IFDG can obtain the most clear improvements. 
The potential reason could be that IFDG not only aligns with the ultimate goal as deduced in Eq.~(\ref{equ:IFDG}), but also carries dense information about traffic congestion of different degrees. To intuitively demonstrate that our IFDG carries more dense information about the intersection than step-wise travel time, we show their associations with other dense rewards, respectively in Fig.~\ref{fig:ifdg_reward} and Fig.~\ref{fig:stt_reward}. We observe that IFDG is positively associated with the other dense rewards while step-wise travel time is not.

\subsubsection{{Spatial-Temporal Augmentation}}
To investigate the individual effect of position encoding and consecutive observation separately, we conduct experiments that concatenate the original observation with either position encoding (P) or consecutive observation (C). As shown in the middle part of Tab.~\ref{tab:ablation}, in comparison to the results of the ``Ideal-Factual Distance Gap" which only uses the original observation, we find that TSC agents can obtain further improvements with either spatial or temporal information. Especially with position encoding (P), TSC agents can achieve consistent gains. However, we also discover that in the case of $\mathcal{D}_\text{NY196}$, the performance degrades with consecutive observation alone while using both information does not witness the same degradation. To this end, we suppose the potential reason is that in a large road network, the location of the intersection plays a critical role in predicting future traffic congestion. And the consecutive observations without the context of the location of the intersection, i.e., Consecutive Observation (C) in Tab.~\ref{tab:ablation}, might be a piece of misleading information. We visualize the curves of future congestion estimation losses with different components in Fig.~\ref{fig:valueloss}. In Fig.~\ref{fig:newyork_valueloss}, the future traffic congestion prediction is worse with only consecutive observations. Moreover, from Fig.~\ref{fig:valueloss}, we can observe that the patterns of estimation losses are in accordance with the results in Tab.~\ref{tab:ablation}, which implies that the average travel time has an association with the accuracy of value estimations. This is because, by using IFDG as a reward, the value is the expectation of the discounted accumulated ideal-factual distance gap, positively associated with the average travel time. This observation also justifies the benefit of our IFDG reward as an unbiased measurement of the average travel time.

\subsubsection{{Non-Local Fusion}}
As shown in Tab.~\ref{tab:ablation}, adding the non-local branch achieves consistent improvement on all tasks. Especially under $\mathcal{D}_\text{NY196}$, non-local information enhancement results in a breakthrough in that the average travel time approaches 800 seconds. To comprehensively analyze the advantages of our non-local branch, we conduct analyses from {three} aspects including \textit{the accuracy of future traffic conditions}, \textit{communication mechanisms} and \textit{{dense communication layers}}.

\paragraph{{The accuracy of future traffic conditions.}} From Fig.~\ref{fig:valueloss}, we observe that, by adding a non-local branch, the value losses of DenseLight obtain further improvements in comparison with others. Especially on tasks $\mathcal{D}_\text{HZ16(f)}$ and $\mathcal{D}_\text{NY196}$, the value losses when using non-local branch drop significantly, which is consistent {with the results  in Tab.~\ref{tab:ablation} (the last row of ``S.T. Aug." part v.s. ``DenseLight")}. Given the positive association between value using IFDG and future average travel time, this observation highlights the importance of aggregating non-local traffic conditions in predicting the long-horizon future traffic condition of each intersection. 

\paragraph{{Non-local communication.}}
{To} highlight the advantage of learning to propagate and integrate information from non-local intersections against manually specified communication among neighboring intersections, we include two sets of experiments that use fixed attention weights, i.e., $W_\text{1-hop}$ and $W_\text{2-hop}$, in the dense communication layer. $W_\text{1-hop}$, for example, assigns non-zero $\frac{1}{|I_\text{1-hop}^i|}$ to the $W_\text{1-hop}^{i, j}, \forall j \in I_\text{1-hop}^i$, where $I_\text{n-hop}^i$ includes intersections that can be reached at most $n$ steps from the intersection $i$. As shown in Tab.~\ref{tab:ablation} {(``1-hop", ``2-hop" and ``DenseLight")}, with the learnable $W$, DenseLight can outperform both $W_\text{1-hop}$ and $W_\text{2-hop}$ consistently.

\paragraph{{The effectiveness of dense communication layer.}}
To investigate the effectiveness, {we modify our DenseLight by} replacing the non-local branch with a transformer encoder~\cite{vaswani2017attention,wang2022synchronous}. The transformer is a well-known architecture to capture the global relationships among language tokens/image patches. From the results in Tab.~\ref{tab:ablation}, DenseLight+Transformer can outperform DenseLight without a non-local branch consistently and has better results against both 1-hop and 2-hop in most cases. This observation again justifies the benefit of aggregating non-local information in solving TSC problems. However, efficient as the transformer is, it is usually over-parameterized (the total size of parameters is about 2.8 MB) and computation-heavy, making RL difficult to improve. Different from the transformer, our non-local branch only uses a linear weighted sum to aggregate non-local information, resulting in a more slim model (the total size of parameters is about 0.2 MB) and an easier RL optimization process.

\section{Conclusion}
In this paper, we propose a novel method DenseLight for a multi-intersection traffic signal control problem (TSC) based on deep reinforcement learning (RL). Specifically, DenseLight optimizes the average travel time of vehicles in the road network under the guidance of an unbiased and dense reward named Ideal-Factual Distance Gap (IFDG) reward and further benefits the future accumulated IFDG modeling by a Non-local enhanced TSC (NL-TSC) agent through spatial-temporal augmented observation and non-local information fusion. We conduct comprehensive experiments on several real-world road networks and various traffic flows, and the results demonstrate consistent performance improvement.
In the future, we will extend our method to learn policies with generalization ability over different road networks.

\section*{Acknowledgments}
This work was supported in part by National Key R\&D Program of China under Grant No.2021ZD0111601, and the Guangdong Basic and Applied Basic Research Foundation (Nos.2023A1515011530, 2021A1515012311, and 2020B1515020048), in part by the National Natural Science Foundation of China (Nos.62002395 and 61976250), in part by the Shenzhen Science and Technology Program (No.JCYJ20220530141211024), in part by the Fundamental Research Funds for the Central Universities under Grant 22lgqb25, and in part by the Guangzhou Science and Technology Planning Project (No.2023A04J2030).

\section*{{Contribution Statement}}
Junfan Lin and Yuying Zhu have equal contributions. Yang Liu is the corresponding author. All of the authors have contributed to the experiments and writing of this work.

\bibliographystyle{named}
\bibliography{ijcai23}

\section*{Appendix}

\section{Notations}
For a better reading, we enumerate several frequently-used (at least twice) symbols in our paper for easy reference, as shown in Tab.~\ref{tab:symb}. And the overall DenseLight algorithm is shown in Alg.~\ref{alg:denselight}.

\begin{table}[t]
\centering 
\footnotesize{
\begin{tabular}{ll}
\toprule
Symbol & Meaning \\ \midrule
$a$ & action/selected traffic phase \\
D & the number of selected phases of an intersection \\
$d$ & $d$-th selected traffic phase \\
$h$ & hidden states of a deep neural network \\
$I$ & all traffic intersections in a road network \\ 
$L$ & all traffic lanes of an intersection \\
$o$ & the output (e.g., action logits or value) of the model \\
$p$ & position encoding \\
$r$ & reward \\
$r_\text{STT}$ & step-wise travel time reward \\
$r_\text{IFDG}$ & ideal-factual distance gap reward \\
$s$ & observation of a traffic intersection \\
T & time in seconds \\
$T_\text{TSC}$ & total duration of a task \\
$T_\text{phase}$ & the duration of a traffic phase\\
$v$/$v(x)$ & speed/speed of a vehicle $x$  (m/s)\\
$W$ & parameters of a Non-local Fusion Layer \\
X & all vehicles of a task \\
$x_e$/$x_l$ & the moment a vehicle $x$ enters/leaves \\
$\pi$ & traffic signal control policy \\
$\theta$ & parameters of traffic signal control policy \\
$\phi$ & parameters of value function \\
$\mathcal{A}$ & advantage function \\
$\mathcal{B}$ & data buffer for training \\
$\mathcal{D}$ & a traffic signal control task \\
$\mathcal{T}$ & dynamics transition mapping of a task \\
$\mathcal{V}$ & value function \\
ConvNet & a stack of $1\times 1$ convolutional layers \\
    \bottomrule
\end{tabular}}
\caption{Symbols used in the paper and their meanings.}
\label{tab:symb}
\end{table}

\section{More Details}
\subsection{Task details}
\begin{figure*}[t]
\centering
\subfloat[\footnotesize{Dongfeng, Jinan, China}]{
	\includegraphics[width=.23\linewidth]{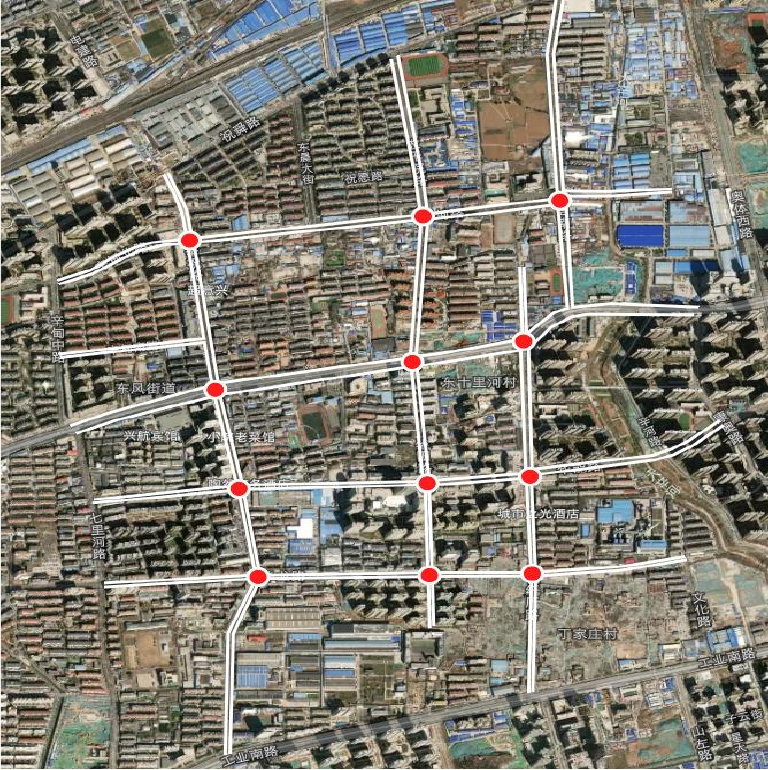}
	\label{fig:jinan_roadnet}
}
\subfloat[\footnotesize{Gudang, Hangzhou, China}]{
	\includegraphics[width=0.23\linewidth]{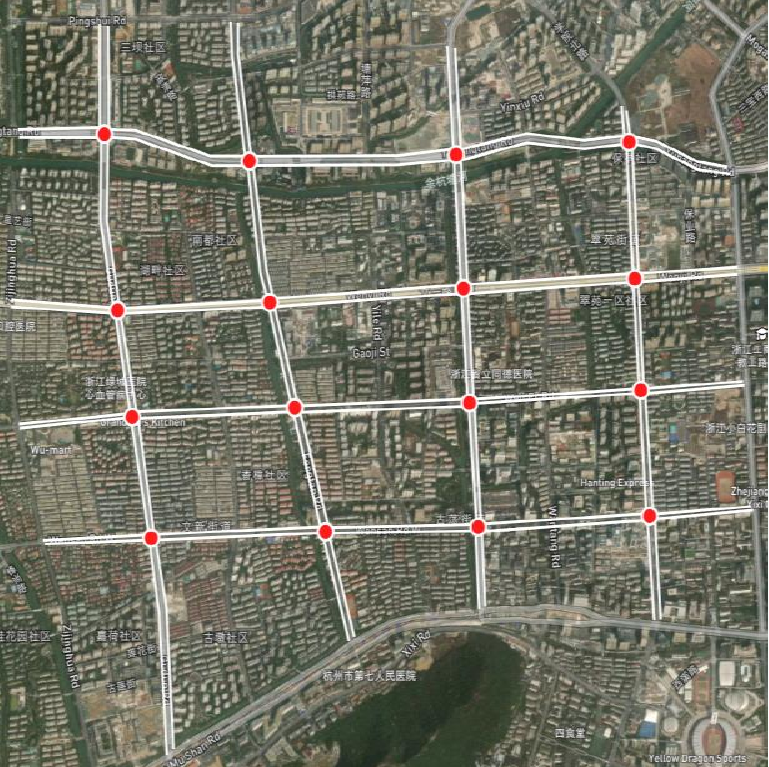}
	\label{fig:hz_roadnet}
}
\subfloat[\footnotesize{Midtown, New York, USA}]{
	\includegraphics[width=0.23\linewidth]{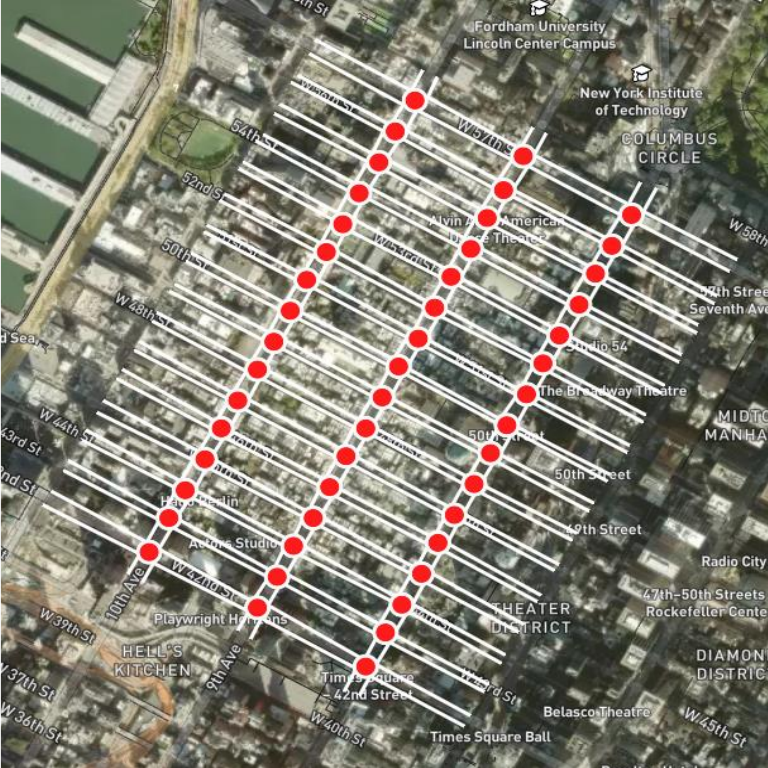}
	\label{fig:nyc48_roadnet}
}
\subfloat[\footnotesize{Upper East, New York, USA}]{
	\includegraphics[width=0.23\linewidth]{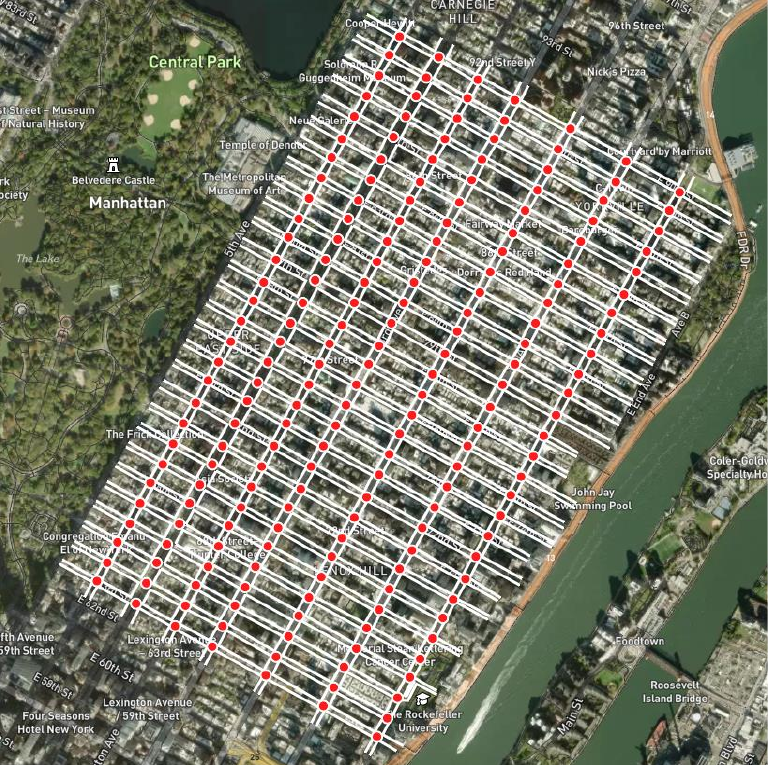}
	\label{fig:nyc196_roadnet}
}
\caption{Area maps of the real-world road networks Jinan12, Hangzhou16, NewYork48 and NewYork196. Red dots are the traffic signals controlled by the agents and the roads in these intersections are marked by white lines.}
\label{fig:roadnet_satellite}
\end{figure*}

\subsubsection{Traffic Signal Control Examples} 

\begin{figure}[!t]
    \centering
    \includegraphics[width=\linewidth]{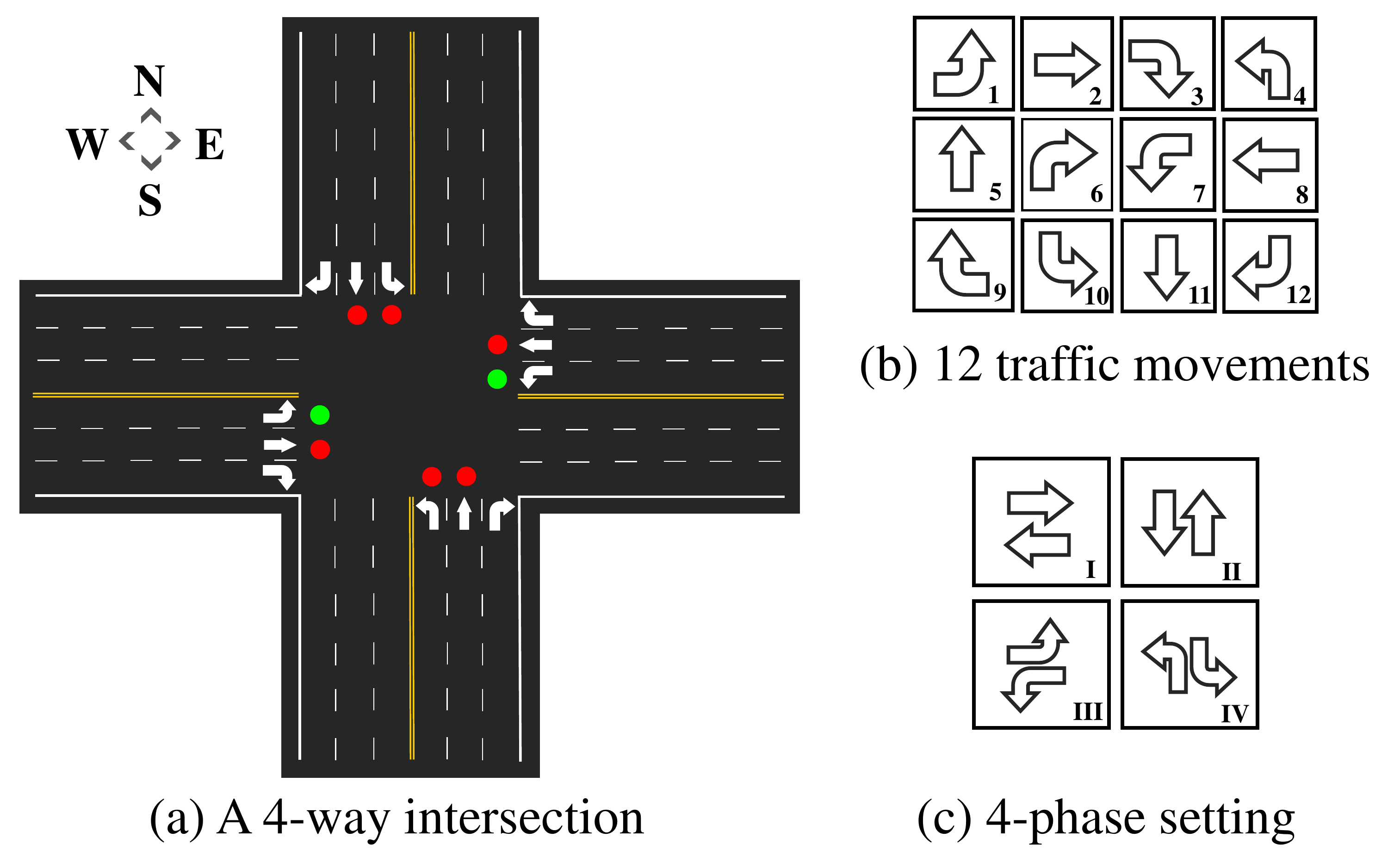}
    \caption{A 4-way intersection structure with its traffic movements and a 4-phase setting. Phase III is set in (a), and traffic movements 1 and 7 are allowed.}
    \label{fig:intersection}
\end{figure}

Several TSC terms are defined in the following with a typical 4-way intersection illustrated in Fig.~\ref{fig:intersection} (a) as an example.

\subsubsection{Traffic Road Networks Examples}

To better understand the road network structures of the datasets, we have depicted their real-world demonstration in Fig.~\ref{fig:roadnet_satellite}.

\subsection{Step-wise Travel Time}

Naturally, it is reasonable to adopt the step-wise travel time for reward design. Formally, for an intersection $i$, the step-wise travel time reward $r_{\text{STT}, d}^i$ of taking action $a_d^i$ at the $d$-th step is formulated as:
\begin{align}
    c_d^i = \sum_{\forall x^i, t_d < x_l^i} x_l^i - \max \{t_d, x_e^i\}, \text{ and } r_{\text{STT}, d}^i = - (c_{d}^i - c_{d+1}^i),
\end{align}
where $c_{d}^i$ represents the sum of the remaining travel time of vehicles that leaves the intersection $i$ after $t_d$, $x_e^i$ and $x_l^i$ denote the moments that vehicle $x$ enters and leaves the intersection $i$, respectively. The accumulation of $r_{\text{STT}, d}^i$ across intersections $I$ results in the opposite of the total travel time:
\begin{align}
    -\sum_{i \in I} \sum_{d=0}^{D-1} r_{\text{STT}, d}^i =  \sum_{i \in I} \sum_{d=0}^{D-1} c_{d}^i - c_{d+1}^i = \sum_{x\in X} x_l - x_e.
\end{align}
However, the step-wise travel time is likely to carry sparse information about the intersection. As formulated in the following, 
\begin{align}
    -r_{\text{STT}, d}^i = &\sum_{\forall x^i, t_d < x_l^i \leq t_{d+1}} (x_l^i - \max \{ t_d,x_e^i\}) \nonumber \\  +&\sum_{\forall x^i, (t_{d+1} < x_l^i) \wedge (x_e^i \leq t_{d+1})} (t_{d+1} - \max \{t_d,x_e^i\}), \label{equ:stt2}
\end{align}
where $r_{\text{STT}, d}^i$ can be divided into two parts in which vehicles leave and stay at the intersection $i$ during $(t_d, t_{d+1}]$, respectively. Normally, there are various reasons for vehicles to stay at the intersection. 
No matter how heavy the congestion of the intersection is, a vehicle $x^i$ that failed to leave the intersection during $(t_d, t_{d+1}]$ results in the same $(t_{d+1} - \max \{t_d,x_e^i\})$. 
In this sense, the second part of $r_{\text{STT}, d}^i$ fails to bring about informative feedback on the current phase selection. Since the duration is usually too short for most vehicles to leave the intersection, which makes the second part in Equ.~(\ref{equ:stt2}) dominates the reward. In this sense, $r_{\text{STT}, d}^i$ carries sparse information and is difficult to reflect the difference among the phases.
In this case, rewards insensitive to the actions are the so-called sparse rewards~\cite{pathak2017curiosity}. Learning RL agent with sparse reward usually results in a locally optimal solution.

\begin{algorithm}[t]
    \caption{DenseLight}
    \label{alg:denselight}
\begin{algorithmic}[1]
    \STATE \textbf{Input}: initial policy parameters $\theta_0$, initial value function parameters $\phi_0$, the number of training epochs $K$
    \STATE \textbf{Output}: $\pi_\theta$
    \FOR{$k = 0,1,2,..., K-1$}
    \STATE Collect set of trajectories ${\mathcal B}_k$ by selecting phase $a_d^i$ by $\pi_{\theta_k}$ using $\hat{s}_d^i$, $\forall d\in [0, D), i\in I$.
    \STATE Calculate $r_{\text{IFDG}, d}^i, \forall d \in [0, D), i \in I$.
    \STATE Compute returns $\eta_d, \forall d\in [0, D)$.
    \STATE Update $\theta$ of policy by Equ.~(\ref{equ:ppoclip}) via Adam gradient ascent.
    \STATE Update $\phi$ of value estimator by Equ.~(\ref{equ:ppovalue}) via Adam gradient ascent.
    \ENDFOR
\end{algorithmic}
\end{algorithm}

\begin{table*}[t]
\newcommand{\tabincell}[2]{\begin{tabular}{@{}#1@{}}#2\end{tabular}}
  \centering
   {
    \begin{tabular}{cccccccc}
    \toprule
     Method & $\mathcal{D}_\text{JN12}$ & $\mathcal{D}_\text{JN12(2)}$ & $\mathcal{D}_\text{JN12(f)}$ & $\mathcal{D}_\text{HZ16}$ & $\mathcal{D}_\text{HZ16(f)}$ & $\mathcal{D}_\text{NY48}$ & $\mathcal{D}_\text{NY196}$\\
    \midrule
    1-hop
    & 228.17 & 216.12 & 240.21 & 249.17 & 288.81 & 157.78 & 926.85 \\
    2-hop
    & 228.23 & 216.31 & 240.36 & 249.10 & 284.52 & 158.18 &  931.62 \\
    DenseLight
    & \textbf{226.97} & {215.82} & {239.58}
    & \textbf{248.43} & \textbf{272.27} 
    & \textbf{156.30} & \textbf{803.42} \\
    DenseLight + Softmax
    & 227.46 & \textbf{215.56} & \textbf{239.47}
    & 248.74 & 273.84 & 159.83 & {846.90} \\
    \bottomrule
    \end{tabular} 
     \caption{The average travel time results of DenseLight using different message communication mechanisms.}
  \label{tab:ablation_for_nonlocal}
  }
\end{table*}

\subsection{Optimization Details}

In RL, optimization methods can be roughly categorized into two branches, i.e., Q-learning~\cite{mnih2015human,watkins1992q} and policy gradient~\cite{sutton1999policy}. Policy gradient methods generally have much smaller value estimation biases than Q-learning-based methods and achieve better performance in practice~\cite{schulman2017proximal}. In this paper, we optimize our DenseLight by a well-known policy gradient algorithm, proximal policy optimization (PPO)~\cite{schulman2017proximal}. We use $\theta_{k}$ and $\phi_{k}$ to denote the parameters of policy $\pi_{\theta}$ and value estimator $\mathcal{V}_\phi$ of PPO at the $k$-th optimization step. For briefness, we used the bold font to indicate the list of items for all intersections $I$. For an example, $\textbf{s}$ represents $[s^1, s^2, ..., s^{|I|}]$. The optimization step is formulated by:
\begin{align}
    \nonumber
    \theta_{k+1} &= \arg \max_{\theta} \frac{1}{|{\mathcal B}_k| D} \sum_{\tau \in {\mathcal B}_k} \sum_{d=0}^{D-1} \min\\ 
    &\left(
            \frac{\pi_{\theta}(\textbf{a}_d|\hat{\textbf{s}}_d)}{\pi_{\theta_k}(\textbf{a}_d|\hat{\textbf{s}}_d)}  \mathcal{A}^{\pi_{\theta_k}}(\hat{\textbf{s}}_d,\textbf{a}_d), 
            \text{B}_\epsilon(\mathcal{A}^{\pi_{\theta_k}}(\hat{\textbf{s}}_d,\textbf{a}_d))
        \right),
    \label{equ:ppoclip} \\
    \phi_{k+1} &= \arg \min_{\phi} \frac{1}{|{\mathcal B}_k| D} \sum_{\tau \in {\mathcal B}_k} \sum_{d=0}^{D-1}\left( \mathcal{V}_{\phi} (\hat{\textbf{s}}_d) - \boldsymbol{\eta}_{d} \right)^2,
    \label{equ:ppovalue}
\end{align}
where $\mathcal{B}$ is the data buffer comprised of several collected episodes $\tau$. $\mathcal{A}^\pi(\textbf{s}, \textbf{a})$ is the advantage function corresponding to a policy $\pi$ measuring the relative gain by taking the actions $\textbf{a}$ with observations $\textbf{s}$. PPO-Clip uses generalized advantage estimator in \cite{schulman2015high}. To prevent the mismatch between the collected episodes and the updated policy, PPO-Clip clips the change of policy within $\epsilon$ by $\text{B}_\epsilon$:
\begin{align}
    \text{B}_\epsilon(\mathcal{A}) = (1 + \mathbbm{1}[\mathcal{A} \geq 0] \epsilon - \mathbbm{1}[\mathcal{A} < 0] \epsilon)\mathcal{A} 
\end{align}

\subsection{Weight visualization}

\begin{figure}[!t]
\centering
\includegraphics[width=\linewidth, clip=True, trim={0, 50, 390, 0}]{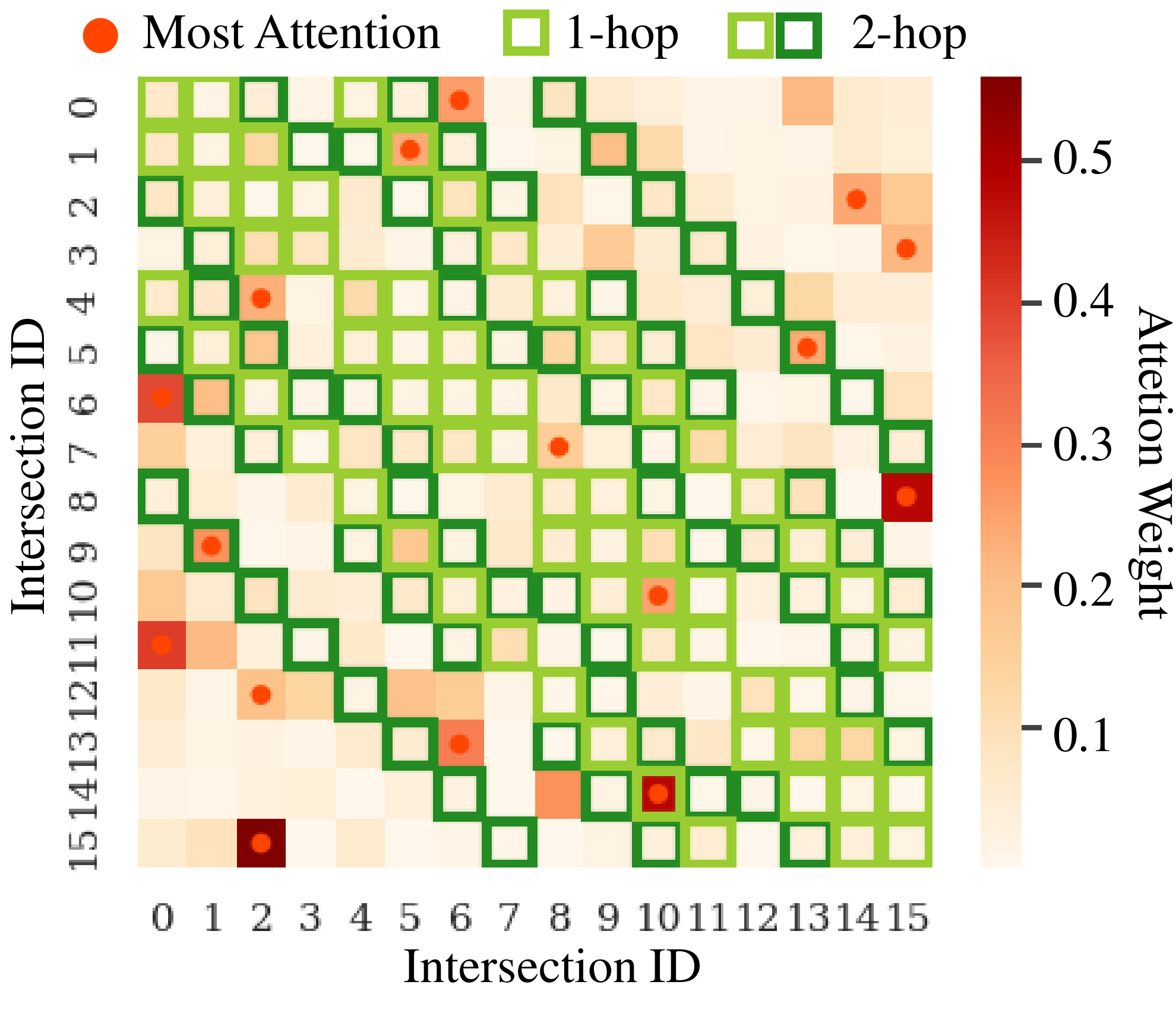}
\caption{Attention weights of each two intersections. The 1-hop and 2-hop neighbors of each intersection (row) are bounded by light green and dark green boxes, respectively. The most attended intersection of each intersection (row) is marked by a red dot.}
\label{fig:w_att}
\end{figure}

To visualize the learned weights between each pair of intersections, we adopt a softmax-version of $W$ which applies softmax operation along the last dimension. As shown in Tab.~\ref{tab:ablation_for_nonlocal}, the softmax-version NL-TSC still outperforms 1-hop and 2-hop. We plot the attention softmax heatmap of the first $W$ of the NL-TSC value estimator in Fig.~\ref{fig:w_att}. From the plot, we can observe that the most attended intersection (hovered with a red dot) of each intersection is not necessarily located in 1-hop (in light-green) or 2-hop (in either light-green or dark-green). According to these observations, we can see that forcing the message to pass locally could hamper the performance of the TSC agent. Interestingly, we also observe that in most cases, 2-hop performs worse than 1-hop, which further implies that the wide-range information does not necessarily contribute to better performance without specifying the importance of different intersections correctly.

\end{document}